\documentclass[journal]{IEEEtran}
\ifCLASSINFOpdf
\else
   \usepackage[dvips]{graphicx}
\fi

\usepackage{graphicx}
\usepackage{graphicx}
\usepackage{comment}
\usepackage{amsmath,amssymb} 
\usepackage{color}
\usepackage{subfigure}
\usepackage{booktabs}
\usepackage{multirow}
\usepackage{threeparttable}
\usepackage{float}

\begin{document}

\title{Learning Person Re-identification Models from Videos with Weak Supervision
}
\author{Xueping Wang, Sujoy Paul, Dripta S. Raychaudhuri, Min Liu, Yaonan Wang and Amit K. Roy-Chowdhury, \IEEEmembership{Fellow, IEEE}
\thanks{Xueping Wang, Min Liu and Yaonan Wang are with the College of Electrical and Information Engineering at Hunan University, Changsha, Hunan, China. 

Amit K. Roy-Chowdhury, Sujoy Paul and Dripta S. Raychaudhuri are with the Department of Electrical and Computer Engineering at the University of California, Riverside.

E-mails: (wang\_xueping@hnu.edu.cn, spaul003@ucr.edu, drayc001@ucr.edu, liu\_min@hnu.edu.cn, yaonan@hnu.edu.cn, amitrc@ee.ucr.edu)

This work was done while Xueping Wang was a visiting student at UC Riverside.}}

\maketitle

\begin{abstract}
Most person re-identification methods, being supervised techniques, suffer from the burden of massive annotation requirement. Unsupervised methods overcome this need for labeled data, but perform poorly compared to the supervised alternatives. In order to cope with this issue, we introduce the problem of learning person re-identification models from videos with weak supervision. The weak nature of the supervision arises from the requirement of video-level labels, {\em i.e.} person identities who appear in the video, in contrast to the more precise frame-level annotations. 
Towards this goal, we propose a multiple instance attention learning framework for person re-identification using such video-level labels.
Specifically, we first cast the video person re-identification task into a multiple instance learning setting, in which person images in a video are collected into a bag. The relations between videos with similar labels can be utilized to identify persons, on top of that, we introduce a co-person attention mechanism which mines the similarity correlations between videos with person identities in common.  
The attention weights are obtained based on all person images instead of person tracklets in a video, making our learned model less affected by noisy annotations. 
Extensive experiments demonstrate the superiority of the proposed method over the related methods on two weakly labeled person re-identification datasets.

\end{abstract}


\IEEEpeerreviewmaketitle

\section{Introduction}
\IEEEPARstart{P}{erson} re-identification (re-id) is a cross-camera instance retrieval problem which aims at searching for persons across multiple non-overlapping cameras \cite{ye2020purifynet,chen2019spatial,chen2019self,chen2020learning,meng2019weakly,yu2020weakly,ye2019dynamic,ye2017dynamic}. 
This problem has attracted extensive research, but most of the existing works focus on supervised learning approaches \cite{chen2019spatial,rao2019learning,sun2018beyond,ye2020purifynet,si2018dual}. While these techniques are extremely effective, they require a substantial amount of annotations which becomes infeasible to obtain for large camera networks.  
Aiming to reduce this huge requirement of labeled data, unsupervised methods have drawn a great deal of attention \cite{lin2019bottom,fan2018unsupervised,qi2019novel,yu2019unsupervised,li2018unsupervised,li2019cross}. However, the performance of these methods is significantly weaker compared to supervised alternatives, as the absence of labels makes it extremely challenging to learn a generalizable model. 

\begin{figure}[]
\centerline{\includegraphics[width=\columnwidth]{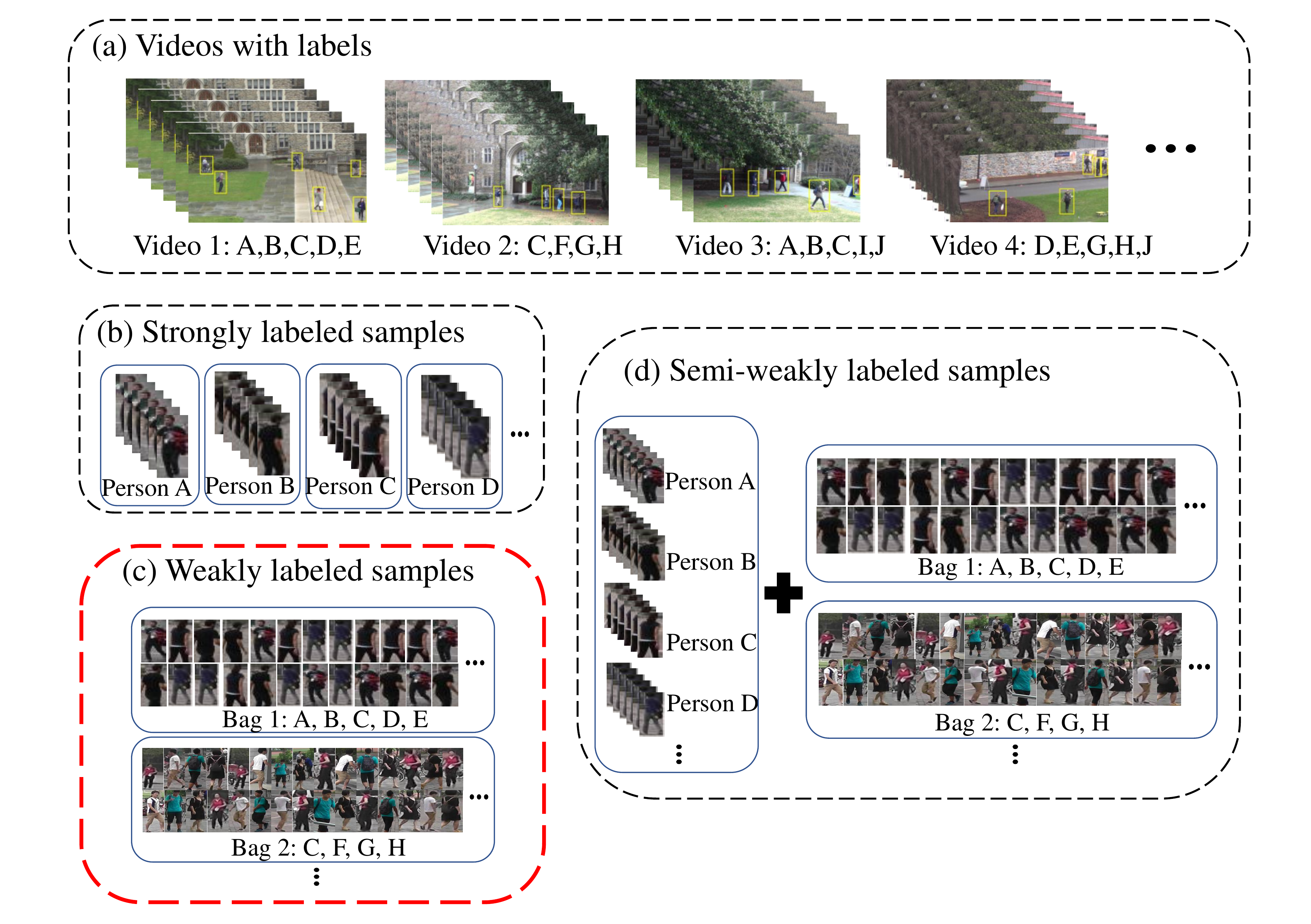}}
\caption{An illustrative example of video person re-id data with multi-level supervisions. (a) shows some raw videos tagged by video-level labels, such as person \{A, B, C, D, E\} for video 1; (b) illustrates the strong labeling setting. The annotators label and associate the person images with the same identity in each video. So, each person image in the video is labeled by their corresponding identity; (c) shows weakly labeled samples {\bf (OURS)}, in which each bag contains all person images obtained in the corresponding video clip and is annotated by the video label without data association and precise data annotations. (d) demonstrates some semi-weakly labeled samples used in \cite {meng2019weakly}, in which the strongly labeled tracklets (one for each identity) in addition to the weakly labeled data are required.
}
\label{fig:1}
\vspace{-0.3cm}
\end{figure}

To bridge this gap in performance, some recent works have focused on the broad area of learning with limited labels. This includes settings such as the one-shot, the active learning and the intra-camera labeling scenarios. The one-shot setting \cite{wu2018exploit,wu2019unsupervised,wu2019progressive,bak2017one} assumes a singular labeled tracklet for each identity along with a large pool of unlabeled tracklets, the active learning strategy \cite{roy2018exploiting,liu2019deep,wang2016human} tries to select the most informative instances for annotation, and the intra-camera setting \cite{zhu2019intra,wang2019weakly} works with labels which are provided only for tracklets within an individual camera view. All of these methods assume smaller proportions of labeling in contrast to the fully supervised setting, but assume \emph{strong labeling} in the form of identity labels similar to the supervised scenario. In this paper, we focus on the problem of learning with \emph{weak labels} - labels which are obtained at a higher level of abstraction, at a much lower cost compared to strong labels. In the context of video person re-id, weak labels correspond to video-level labels instead of the more specific labels for each image/tracklet within a video. 

\begin{figure*}[t]
\centerline{\includegraphics[width=1.99\columnwidth]{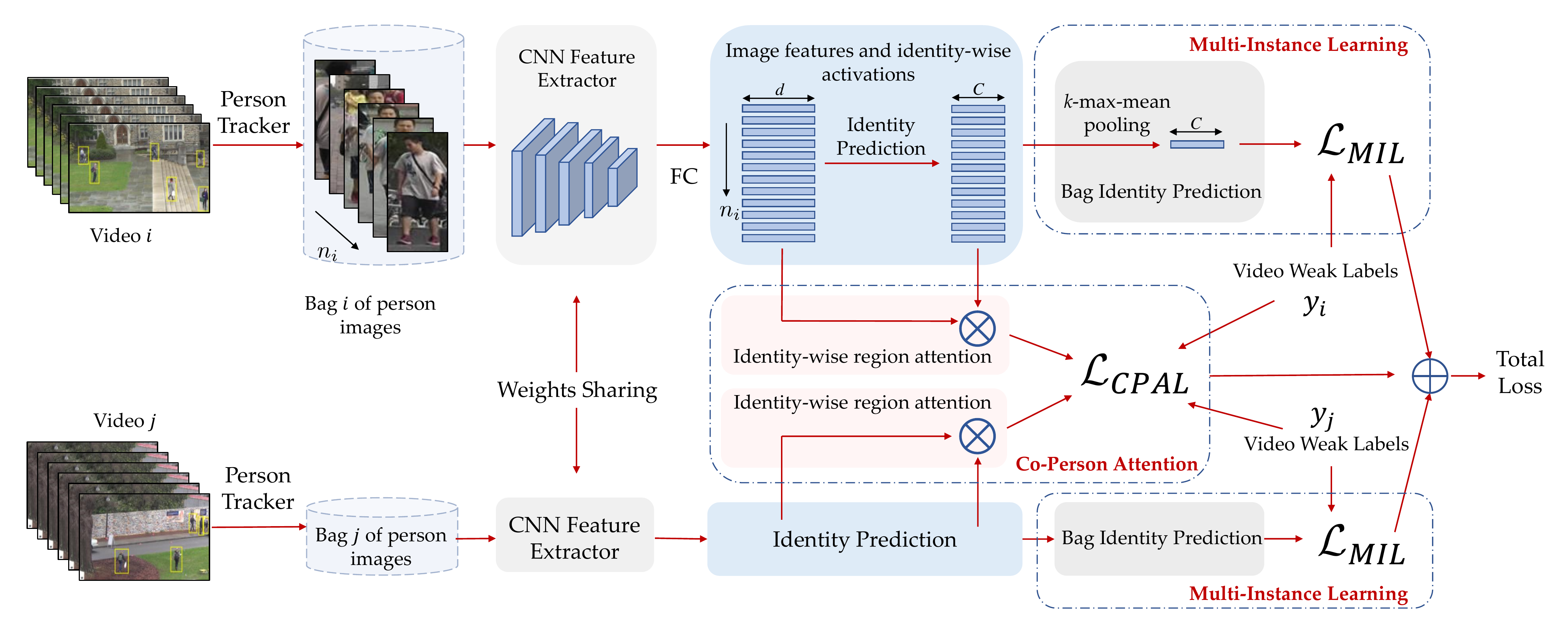}}
\caption{A brief illustration of our proposed multiple instance attention learning framework for video person re-id with weak supervision. For each video, we group all person images obtained by pedestrian detection and tracking algorithms in a bag and use it as the inputs of our framework. The bags are passed through a backbone CNN to extract features for each person image. Furthermore, a fully connected (FC) layer and an identity projection layer are used to obtain identity-wise activations. On top of that, the MIL loss based on \emph k-max-mean-pooling strategy is calculated for each video. For a pair of videos $(i,j)$ with common person identities, we compute the CPAL loss by using high and low attention region for the common identity. Finally, the model is optimized by jointly minimizing the two loss functions.
}
\label{fig:2}
\end{figure*}

To illustrate this further, consider Figure \ref{fig:1} which shows some video clips which are annotated with the video-level labesls, such as video 1 with \{A, B, C, D, E\}. This indicates that Person A, B, C, D and E appear in this clip. By using pedestrian detection and tracking algorithms \cite{lin2017feature,ren2015faster,jin2019multi}, we can obtain the person images (tracklets) for this video clip, but can make no direct correspondence between each image (tracklet) and identity due to the weak nature of our labels. Specifically, we group all person images obtained in one video clip into a bag and tag it with the video label as shown in Figure 1(c). On the contrary, strong supervision requires identity labels for each image (tracklet) in a video clip and thus, annotation is a more tedious procedure compared to our setting.
Thus, in weakly labeled person re-id data, we are given bags, with each such bag containing all person images in a video and the video's label; our goal is to train a person re-id model using these bags which can perform retrieval during test time at two different levels of granularity. The first level of granularity, which we define as \emph{Coarse-Grained Re-id}, involves retrieving the videos (bags) that a given target person appears in. The second level entails finding the exact tracklets with the same identity as the target person in all obtained gallery tracklets - this is defined as \emph{Fine-Grained Re-id}. Moreover, we also consider a more practical scenario where the weak labels are not reliable - the annotators may not tag the video clip accurately.

In order to achieve this goal, we propose a multiple instance attention learning framework for video person re-id which utilizes pairwise bag similarity constraints via a novel co-person attention mechanism. Specifically, we first cast the video person re-id task into a multiple instance learning (MIL) problem which is a general idea that used to solve weakly-supervised problems \cite{meng2019weakly,bilen2016weakly}, however, in this paper, a novel {\em k}-max-mean-pooling strategy is used to obtain a probability mass function over all person identities for each bag and the cross-entropy between the estimated distribution and the ground truth identity labels for each bag is calculated to optimize our model. The MIL considers each bag in isolation and does not consider the correlations between bags. We address this by introducing the Co-Person Attention Loss (CPAL), which is based on the motivation that a pair of bags having at least one person identity {\em e.g. Person A} in common should have similar features for images which correspond to that identity ({\em A}). Also, the features from one bag corresponding to {\em A} should be different from features of the other bag (of the pair) not corresponding to {\em A}. We jointly minimize these two complementary loss functions to learn our multiple instance attention learning framework for video person re-id as shown in Figure \ref{fig:2}. 

To the best of our knowledge, this is the first work in video person re-id which solely utilizes the concept of weak supervision. A recent work \cite{meng2019weakly} presents a weakly supervised framework to learn re-id models from videos. However, they require strong labels, one for each identity, in addition to the weak labels, resulting in a semi-weak supervision setting. In contrast, our setting is much more practical forgoing the need for \emph{any} strong supervision. A more detailed discussion on this matter is presented in Section \ref{sec:cv_miml}, where we empirically evaluate the dependence of \cite{meng2019weakly} on the strong labels and demonstrate the superior performance of our framework.


\emph{Main contributions.} The contributions of our work are as follows: 
\begin{itemize}
    \item[$\bullet$] We  introduce  the  problem  of  learning  a re-id model from videos with weakly  labeled  data and propose a multiple instance attention learning framework to address this task.
    \item[$\bullet$] By exploiting the underlying characteristics of weakly labeled person re-id data, we present a new co-person attention mechanism to utilize the similarity relationships between videos with common person identities.
    \item[$\bullet$] We conduct extensive experiments on two weakly labeled datasets and demonstrate the superiority of our method on coarse and fine-grained person re-id tasks. We also validate that the proposed method is promising even when the weak labels are not reliable. 
\end{itemize}

\section{Related Works}
Existing person re-id works can be summarized into three categories, such as learning from strongly labeled data (supervised and semi-supervised), learning from unlabeled data (unsupervised) and learning from weakly labeled data (weakly supervised) depending on the level of supervision. This section briefly reviews some person re-id works, which are related with this work.

{\bf Learning from strongly labeled data.} Most studies for person re-id are supervised learning-based methods and require the fully labeled data \cite{chen2018improving,sun2018beyond,chen2019spatial,rao2019learning,si2018dual,zheng2016mars,chen2018video}, {\em i.e.}, the identity labels of all the images/tracklets from multiple cross-view cameras. These fully supervised methods have led to impressive progress in the field of re-id; however, it is impractical to annotate very large-scale surveillance videos due to the dramatically increasing annotation cost.

To reduce annotation cost, some recent works have focused on the broad area of learning with limited labels, such as the one-shot settings \cite{wu2018exploit,wu2019unsupervised,wu2019progressive,bak2017one}, the active learning strategy \cite{roy2018exploiting,liu2019deep,wang2016human} and the intra-camera labeling scenarios \cite{zhu2019intra,wang2019weakly}. All of these methods assume smaller proportions of labeling in contrast to the fully supervised setting, but assume strong labeling in the form of identity labels similar to the supervised scenario.

{\bf Learning from unlabeled data.} Researchers developed some unsupervised learning-based person re-id models \cite{lin2019bottom,fan2018unsupervised,qi2019novel,yu2019unsupervised,li2018unsupervised,li2019cross} that do not require any person identity information. 
Most of these methods follow a similar principle - alternatively assigning pseudo labels to unlabeled data with high confidence and updating model using these pseudo-labeled data. It is easy to adapt this procedure to large-scale person re-id task since the unlabeled data can be captured automatically by camera networks.
However, most of these approaches perform weaker than those supervised alternatives due to lacking the efficient supervision.

{\bf Learning from weakly labeled data.} 
The problem of learning from weakly labeled data has been addressed in several computer vision tasks, including object detection \cite{heidarivincheh2019weakly,yu2019temporal,bilen2016weakly}, segmentation \cite{khoreva2017simple,ahn2018learning}, text and video moment retrieval \cite{mithun2019weakly}, activity classification and localization \cite{chen2017attending,paul2018w,nguyen2019weakly}, video captioning \cite{shen2017weakly} and summarization \cite{panda2017weakly,cai2018weakly}. 
There are three weakly supervised person re-id models have been proposed. Wang {\em et al.} introduced a differentiable graphical model \cite{wang2019weakly} to capture the dependencies from all images in a bag and generate a reliable pseudo label for each person image.
Yu {\em et al.} introduced the weakly supervised feature drift regularization \cite{yu2020weakly} which employs the state information as weak supervision to iteratively refine pseudo labels for improving the feature invariance against distractive states.
Meng {\em et al.} proposed a cross-view multiple instance multiple label learning method \cite{meng2019weakly} that exploits similar instances within a bag for intra-bag alignment and mine potential matched instances between bags. However, our weak labeling setting is more practical than these three works for video person re-id. First, we  do  not  require  any strongly labeled tracklets and state information for model training. 
Second, we consider a scenario that the weak labels are not reliable in training data. 

Our task of learning person re-id models from videos with weak supervision is also related to the problem of person search \cite{yan2019learning,xiao2019ian,han2019re} whose objective is to simultaneously localize and recognize a person from raw images. The difference lies in the annotation requirement for training - the person search methods assume large amounts of manually annotated bounding boxes for model training. Thus, these approaches utilize strong supervision in contrast to our weak supervision.

\section{Methodology}
In this section, we present our proposed multiple instance attention learning framework for video person re-id. We first present an identity projection layer we use to obtain the identity-wise activations for input person images in one bag. Thereafter, two learning tasks: multi-instance learning and co-person attention mechanism are introduced and jointly optimized to learn our model. The overview of our proposed method is shown in Figure \ref{fig:2} and it may be noted that only the video-level labels of training data are required for model training. Before going into the details of our multiple instance attention learning framework, let us first compare the annotation cost between weakly labeled and strongly labeled video person re-id data, and then define the notations and problem statement formally.
\subsection{Annotation Cost}
We focus on person re-id in videos, where labels can be collected in two ways:
\begin{itemize}
    \item[$\bullet$]\emph {Perfect tracklets}: The annotators label each person in each video frame with identities and associate persons with the same identity (DukeMTMC-VideoReID \cite{wu2018exploit}). Then, the tracklets are perfect and one tracklet contains one person identity. However, they are more time-consuming than ours which requires only video-level labels.
    \item[$\bullet$]\emph{Imperfect tracklets}: The tracklets are obtained automatically by pedestrian detection and tracking algorithms \cite{lin2017feature,ren2015faster,jin2019multi} (MARS \cite{zheng2016mars}). They are bound to have errors of different kinds, like wrong associations, missed detection, etc. Thus, human intervention is required to segregate individual tracklets into the person identities.
\end{itemize}

Our method uses only video-level annotations, reducing the labeling efforts in both the above cases. We put all person images in a video to a bag and label the bag with the video-level labels obtained from annotators. We develop our algorithm without any idea of the tracklets, but rather a bag of images. Further, we do not use any intra-tracklet loss, as one tracklet can have multiple persons in case of imperfect tracking. Table \ref{table:coarse-grained re-id} and Table \ref{table:fine-grained re-id} show our method is robust against the missing annotation scenario where a person might be there in the video, but not labeled by annotators. Hence, our framework has remarkable real-world value where intra-camera tracking is almost surely to happen with an automated software and will be prone to errors.

Next, we present an approximate analysis of the reduction in annotation cost by utilizing weak supervision. Assume that the cost to label a person in an image is $b$. Also, let the average number of persons per image be $p$ and the average number of frames per video be $f$. The total number of videos from all cameras is $n$. So, the annotation cost for strong supervision is $fpnb$. Now, let the cost for labeling a video with video-level labels be $b'$, where $b'<<b$.  Thus, the annotation cost for weak supervision amounts to $nb'$. This results in an improvement in the annotation efficiency by $fpb/b'\times 100\%$.

\subsection{Problem Statement}
Assume that we have {\em C } known identities that appear in $N$ video clips. In our weakly labeling settings, each video clip is conceptualized as a bag of person images detected in the video, and assigned a label vector indicating which identities appear in the bag. 
Therefore, the training set can be denoted as $\mathcal D=\{(\mathcal X_i,y_i)|i=1,...,N\}$, where $\mathcal X_i=\{I_i^1,I_i^2,...,I_i^{n_i}\}$ is the $i$th bag (video clip) containing $n_i$ person images. Using some feature extractors, we can obtain the corresponding feature representations for these images, which we stack in the form of a feature matrix $X_i\in \mathbb R^{d\times n_i}$;
$y_i=\{y_i^1,y_i^2,...,y_i^C\}\in\{0,1\}^C$ is the label vector of bag {\em i} containing {\em C} identity labels, in which $y_i^c=1$ if the $c$th identity is tagged for $\mathcal X_i$ (person {\em c} appears in video {\em i}) and $y_i^c=0$ otherwise. For the testing probe set, each query is composed of a set of detected images with the same person identity (a person tracklet) in a video clip. 
We define two different settings for the testing gallery set as follows:
\begin{itemize}
    \item[$\bullet$] {\bf Coarse-grained person re-id} tries to retrieve the videos that the given target person appears in. The testing gallery set should have the same settings as the training set - each testing gallery sample is a bag with one or multiple persons.
    \item[$\bullet$] {\bf Fine-grained person re-id} aims at finding the exact tracklets with the same identity as the target person among all obtained tracklets. It has the same goal as the general video person re-id - each gallery sample is a tracklet with a singular person identity. 
\end{itemize}

\subsection{Multiple Instance Attention Learning for Person Re-id}



\subsubsection{Identity Space Projection} 
In our work, feature representation $X_i$ is used to identify person identities in bag $i$. We project $X_i$ to the identity space ($\mathbb R^{C}$, {\em C} is the number of person identities in training set). 
Thereafter, the identity-wise activations for bag {\em i} can be represented as follows:
\begin{equation}
    \mathcal {W}_i=f(X_i;\theta)
\end{equation}
where $f(\cdot;\theta)$ is a $C$ dimensional fully connected layer. $\mathcal W_i\in\mathbb R^{C\times n_i}$ is an identity-wise activation matrix.
These identity-wise activations represent the possibility that each person image in a bag is predicted to a certain identity.

\subsubsection{Multiple Instance Learning} In weakly labeled person re-id data, each bag contains multiple instances of person images with person identities. So the video person re-id task can be turned into a multiple instance learning problem. In MIL, the estimated label distribution for each bag is expected to eventually approximate the ground truth weak label (video label); thus, we need to represent each bag using a single confidence score per identity. 
In our case, for a given bag, we compute the activation score corresponding to a particular identity as the average of top {\em k} largest activations for that identity ({\em k}-max-mean-pooling strategy). 
For example, the identity-$j$ confidence probability for the bag $i$ can be represented as,

\begin{equation}
    p_i^j=\frac{1}{k}\textup{topk}(\mathcal W_i[j,:])
\end{equation}
where topk is an operation that selects the top {\em k} largest activations for a particular identity. $\mathcal W_i[j,:]$ denotes the activation score corresponding to identity {\em j} for all person images in bag {\em i}. 
Thereafter, a softmax function is applied to compute the probability mass function (pmf) over all the identities for bag {\em i} as follows, $\hat y_i^j=\frac{\exp(p_i^j)}{\sum\limits_{k=1}^{C}\exp (p_i^k)}$.
The MIL loss is the cross-entropy between the predicted pmf $\hat y_i$ and the normalized ground-truth $y_i$, which can then be represented as follows,

\begin{figure}
\centerline{\includegraphics[width=\columnwidth]{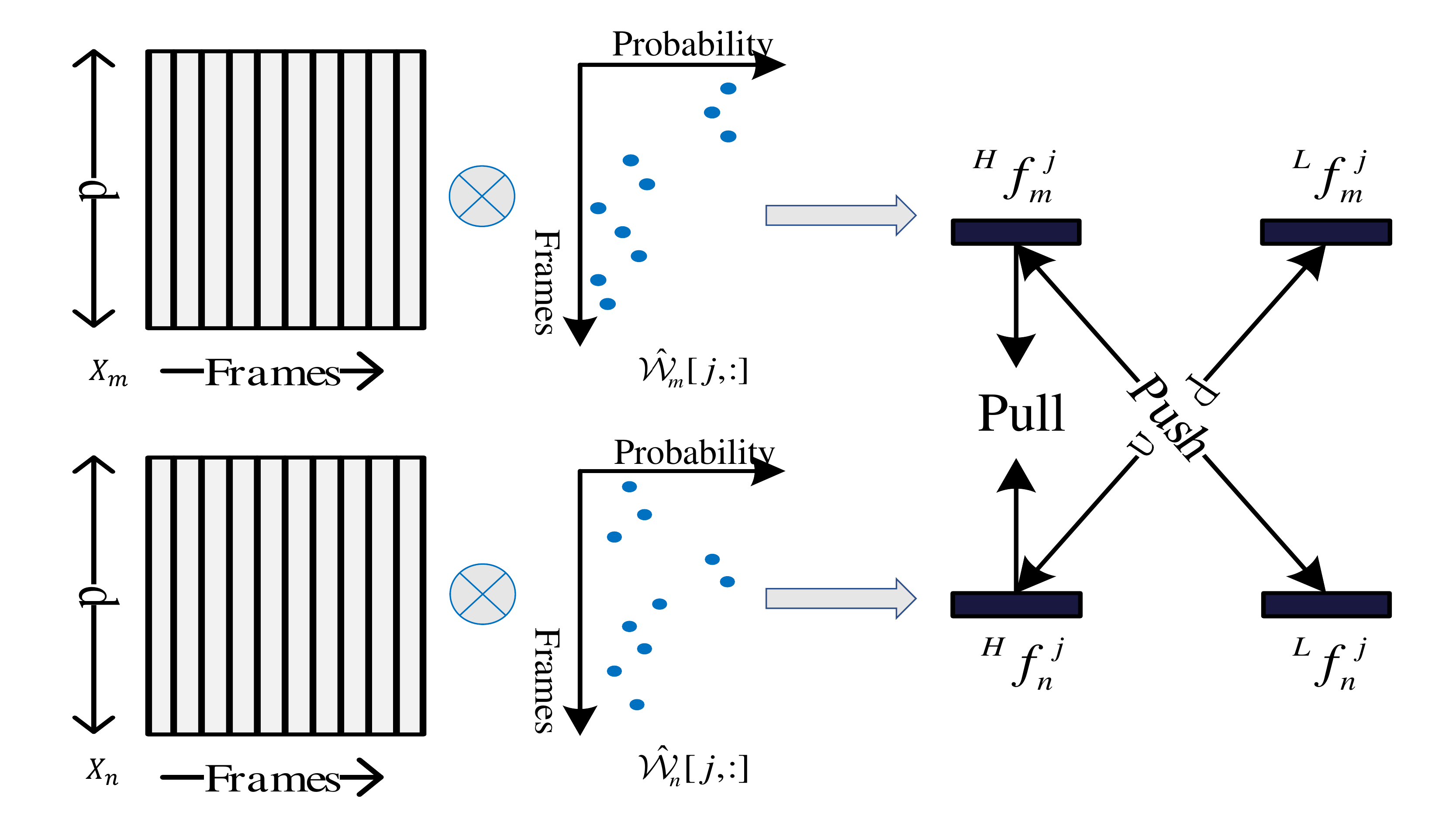}}
\caption{This figure illustrates the procedure of the co-person attention mechanism.  We assume that bag {\em m} and {\em n} have person identity {\em j} in common. We first obtain the feature representations $X_m$ and $X_n$, and identity-{\em j} activation vectors $\hat{\mathcal W}_m[j,:]$ and $\hat{\mathcal W}_n[j,:]$ by passing the bags through our model. Thereafter, high and low identity-{\em j} attention features $^Hf_m^j$ and $^Lf_m^j$ can be obtained for each bag.
Finally, we want the features with high identity-{\em j} attention region to be close to each other, otherwise push them to be away from each other.
}
\label{fig:3}
\end{figure}

\begin{equation}
    \mathcal L_{MIL}=\frac{1}{N_b}\sum\limits_{i=1}^{N_b}\sum\limits_{j=1}^{C}-y_i^j\log(\hat y_i^j)
\end{equation}
where $y_i$ is the normalized ground truth label vector and $N_b$ is the size of training batch.
The MIL only considers each bag in isolation. Next, we present a Co-Person Attention Mechanism for mining the potential relationships between bags. 

\subsubsection{Co-Person Attention Mechanism}
In a network of cameras, the same person may appear at different times and different cameras, so there may be multiple video clips (bags) containing common person identities. That motivates us to explore the similarity correlations between bags. Specifically, for those bags with at least one person identity in common, we may want the following properties in the learned feature representations: first, a pair of bags with {\em Person j} in common should have similar feature representations in the portions of the bag where the {\em Person j} appears in; 
second, for the same bag pair, feature representation of the portion where {\em Person j} occurs in one bag should be different from that of the other bag where {\em Person j} does not occur.

We introduce Co-Person Attention Mechanism to integrate the desired properties into the learned feature representations. In the weakly labeled data, we do not have frame-wise labels, so the identity-wise activation matrix obtained in Equation 1 is employed to identify the required person identity portions. 
Specifically, for bag {\em i}, we normalize the bag identity-wise activation matrix $\mathcal W_i$ along the frame index using softmax function as follows:
\begin{equation}
    \hat{\mathcal W_i}[j,t]=\frac{\exp(\mathcal W_i[j,t])}{\sum_{t^{'}=1}^{n_i}\exp(\mathcal W_i[j,t^{'}])}
\end{equation}
Here {\em t} indicates the indexes of person images in bag {\em i} and $j\in\{1,2,...,C\}$ denotes person identity. $\hat{\mathcal W}_i$ could be referred as {\em identity attention}, because it indicates the probability that each person image in a bag is predicted to a certain identity.
Specifically, a high value of attention for a particular identity indicates its high occurrence-probability of that identity. 
Under the guidance of the identity attention, we can define the identity-wise feature representations of regions with high and low identity attention for a bag as follows:

\begin{equation}
\left\{ 
\begin{array}{lr}
    ^H f_i^j=X_i\hat{\mathcal W}_i[j,:]^T, & \\
    ^L f_i^j=\frac{1}{n_i-1}X_i(\boldsymbol{1}-\hat{\mathcal W}_i[j,:]^T)&
\end{array}
\right.
\end{equation}
where $^Hf_i^j, ^Lf_i^j\in\mathbb R^{d}$ represent the aggregated feature representations of bag {\em i} with high and low identity-{\em j} attention region, respectively. 
It may be noted that in Equation 5, the low attention feature is not defined if a bag contains only one person identity and the number of person images is 1, {\em i.e.} $n_i = 1$.
This is also conceptually valid and in such cases, we cannot compute the CPAL loss. 

We use ranking hinge loss to enforce the two properties discussed above. Given a pair of bags {\it m} and {\it n} with person identity {\em j} in common, the co-person attention loss function may be represented as follows:
\begin{equation}
\begin{aligned}
    &&\mathcal L_{m,n}^j=\frac{1}{2}\{\max(0,s(^Hf_m^j,^Hf_n^j)-s(^Hf_m^j,^Lf_n^j)+\delta)\\
    &&+\max(0,s(^Hf_m^j,^Hf_n^j)-s(^Lf_m^j,^Hf_n^j)+\delta)\}
\end{aligned}
\end{equation}
where $\delta=0.5$ is the margin parameter in our experiment. $s(\cdot,\cdot)$ denotes the cosine similarity between two feature vectors. The two terms in the loss function are equivalent in meaning, and they represent that the features with high identity attention region in both the bags should be more similar than the high attention region feature in one bag and the low attention region feature in the other bag as shown in Figure \ref{fig:3}.

The total CPAL loss for the entire training set may be represented as follows:
\begin{equation}
    \mathcal L_{\textit {CPAL}}=\frac{1}{C}\sum_{j=1}^{C}\frac{1}{\binom{|\mathcal S^j|}{2}}\sum_{m,n\in \mathcal S^j}\mathcal L_{m,n}^j
\end{equation}
where $\mathcal S^j$ is a set that contains all bags with person identity {\em j} as one of its labels. $\binom{|\mathcal S^j|}{2}=\frac{|\mathcal S^j| \cdot (|\mathcal S^j|-1)}{2}$. $m,n$ are indexes of bags.

\subsubsection{Optimization}
The MIL considers each bag in isolation but ignores the correlations between bags, and CPAL mines the similarity correlations between bags. Obviously, they are complementary. So, we jointly minimize these two complementary loss functions to learn our multiple instance attention learning framework for person re-id. It can be represented as follows:

\begin{equation}
    \mathcal L=\lambda\mathcal L_{MIL}+(1-\lambda)\mathcal L_{CPAL}
\end{equation}
where $\lambda$ is a hyper-parameter that controls contribution of $\mathcal L_{MIL}$ and $\mathcal L_{CPAL}$ for model learning. In Section \ref{sec:lambda}, we discuss the contributions of each part for recognition performance. 

\subsection{Coarse and Fine-Grained Person Re-id}
In the testing phase, each query is composed of a set of detected images in a bag with the same person identity (a person tracklet). Following  our  goals, we  have  two  different  settings  for  testing  gallery set.

Coarse-Grained Person Re-id finds the bags (videos) that the target person appears in. So, the testing gallery set is formed in the same manner as the training set.
We define the distance between probe and gallery bags using the minimum distance between average pooling feature of the probe bag and frame features in the gallery bag. Specifically, we use average pooling feature $x_p$ to represent bag {\em p} in the testing probe set and $x_{g,r}$ denotes the feature of $r$th frame in $g$th testing gallery bag. Then, the distance between the bag $p$ and bag $g$ may be represented as follows:

\begin{equation}
    D(p,g)=\min\{d(x_p,x_{g,1}),d(x_p,x_{g,2}),...,d(x_p,x_{g,n_g})\}
\end{equation}
where $d(\cdot,\cdot)$ is the Euclidean distance operator. $n_g$ is the number of person images in bag $g$.

Fine-Grained Person Re-id finds the tracklets with the same identity  as  the  target  person. This goal is same as the general video person re-id, so testing gallery samples are all person tracklets. We evaluate the fine-grained person re-id performance following the general person re-id setting.

\section{Experiments}

\begin{table*}
\caption{Detailed information of two weakly labeled person re-id datasets.}
\centering
\begin{threeparttable}
\setlength{\tabcolsep}{2mm}
\begin{tabular}{lllllllllll}
\toprule
\multirow{3}{*}{Dataset} & \multirow{3}{*}{Settings} & \multicolumn{3}{c}{Training Set}                                             & \multicolumn{6}{c}{Testing Set}                                    \\ \cline{3-11} 
                         &                           & \multirow{2}{*}{IDs} & \multirow{2}{*}{Tracks} & \multirow{2}{*}{Bags} & \multicolumn{3}{c}{Probe Set} & \multicolumn{3}{c}{Gallery Set} \\ \cline{6-11} 
                         &                           &                        &                           &                         & IDs    & Tracks   & Bags   & IDs   & Tracks   & Bags   \\ \hline
\multirow{2}{*}{WL-MARS} & Coarse                    & 625                    & -                         & 2081                    & 626      & -          & 626      & 634     & -          & 1867     \\ 
                         & Fine                      & 625                    & -                         & 2081                    & 626      & 1980       & -        & 636     & 12180      & -        \\ \hline
\multirow{2}{*}{WL-DukeV}                 & Coarse                    & 702                    & -                         & 3842                    & 702      & -          & 702      & 1110    & -          & 483      \\ 
                         & Fine                      & 702                    & -                         & 3842                    & 702      & 702        & -        & 1110    & 2636       & -        \\ \bottomrule
\end{tabular}
IDs, Tracks and Bags denote the number of identities, tracklets and bags. Coarse and Fine represent coarse-grained person re-id and fine-grained setting.
\end{threeparttable}
\label{table:dataset}
\end{table*}

\begin{table*}[t]
\caption{Coarse-grained person re-id performance comparisons. $\downarrow$ represents the decreased recognition performance compared to perfect annotation.}
\centering
\setlength{\tabcolsep}{2mm}{
\begin{tabular}{lllllllll}

\toprule
\multicolumn{1}{c}{\multirow{2}{*}{Methods}} & \multicolumn{4}{c}{WL-MARS}               & \multicolumn{4}{c}{WL-DukeV}              \\ \cmidrule{2-9} 
                         & R-1 & R-5 & R-10 & mAP  & R-1 & R-5 & R-10 & mAP  \\ \midrule
WSDDN \cite{bilen2016weakly}                     & 63.4   & 81.9   & 86.6    &30.3 &72.4   & 89.6     & 93.6   & 62.2   \\
HSLR \cite{dong2019single}                    & 69.6   & 85.9   & 89.8    & 35.4 & 77.5   & 93.0     & 95.2    & 66.0   \\ 
SSLR \cite{dong2019single}                    & 66.6   & 82.7   & 86.6    & 31.8 & 76.2   & 90.5   & 93.6    & 64.2 \\ 
MIL &73.2    & 89.9   & 93.3    & 41.3       &80.8    &93.4    & 95.6    & 69.1\\
OURS (MIL+CPAL)                      & \bf{78.6}   & \bf{90.1}   & \bf{93.9}      & \bf{47.1} & \bf{82.6}   & \bf{93.6}   & \bf{95.6}    & \bf{72.1} \\ 

OURS*                     & 78.1$\downarrow$\tiny{0.5}   & 88.3$\downarrow$\tiny{1.8}  &91.5$\downarrow$\tiny{2.4}   & 42.7$\downarrow$\tiny{4.4} & 79.3$\downarrow$\tiny{3.3}   & 92.7$\downarrow$\tiny{0.9}   &95.4$\downarrow$\tiny{0.2}    & 68.3$\downarrow$\tiny{3.8} \\ 

\bottomrule
\multicolumn{9}{l}{OURS* represents the proposed method under missing annotation.}                   
\end{tabular}}
\label{table:coarse-grained re-id}
\end{table*}

\subsection{Datasets and Settings}
\subsubsection{Weakly Labeled Datasets} We conduct experiments on two weakly labeled person re-id datasets - Weakly Labeled MARS (WL-MARS) dataset and Weakly Labeled DukeMTMC-VideoReID (WL-DukeV) dataset. These two weakly labeled datasets are based on the existing video-based person re-id datasets - MARS \cite{zheng2016mars} and DukeMTMC-VideoReID \cite{wu2018exploit} datasets, respectively. They are formed as follows: first, 3 - 6 tracklets from the same camera are randomly selected to form a bag; thereafter, we tag it with the set of tracklet labels. It may be noted that only bag-level labels are available and the specific label of each individual is unknown. More detailed information of these two weakly labeled datasets are shown in Table \ref{table:dataset}.

We also consider a more practical scenario that the annotator may miss some labels for a video clip, namely, \emph {missing annotation}. For example, one person may only appear for a short time and missed by the annotator. It will lead to a situation that weak labels are not reliable. To simulate this circumstance, for each weakly labeled bag, we randomly add 3 - 6 short tracklets with different identities into it and each tracklet contains 5 - 30 person images. So, the new bags will contain the original person images and the new added ones, but the labels are still the original bag labels. In Section \ref{sec:sota}, we evaluate the proposed method under this situation.

\begin{table*}[t]
\centering
\caption{Fine-grained person re-id performance comparisons. $\downarrow$ represents the decreased recognition performance compared to perfect annotation.}

\begin{threeparttable}
\setlength{\tabcolsep}{1.2mm}
\begin{tabular}{llllllllll}
\toprule
\multirow{2}{*}{Settings} &\multicolumn{1}{c}{\multirow{2}{*}{Methods}}                                                     & \multicolumn{4}{c}{WL-MARS}             & \multicolumn{4}{c}{WL-DukeV}                                   \\ \cline{3-10} 
\multicolumn{1}{c}{}                         &                                                                               & R-1  & R-5 & R-10  & mAP  & R-1 & R-5 & R-10 & \multicolumn{1}{l}{mAP}  \\ \hline
\multirow{5}{*}{\begin{tabular}[c]{@{}l@{}}Weak sup.\end{tabular}} 
&WSDDN \cite{bilen2016weakly} &59.2 &76.4 &82.4 &41.7 &65.4 &84.0 &90.2 &60.7 \\
&HSLR \cite{dong2019single}                                          & 56.4 & 72.6   & 78.3    & 35.8 & 61.7   & 79.8   & 85.0      & 54.7                      \\ 
                                           &SSLR \cite{dong2019single}                                                                            & 51.9 & 69.3   & 75.7    & 31.2 & 56.3   & 76.1   & 83.0    & \multicolumn{1}{l}{50.0}   \\
                                           & MIL &63.6  &79.1 & 84.2 &43.7 &69.1 &83.3 &89.5 &62.0\\
                                           &OURS (MIL+CPAL)                                                                              &\bf{65.0}   & \bf{81.5}   & \bf{86.1}    & \bf{46.0}   & 70.5   & \bf{87.2}   & \bf{92.2}    & \multicolumn{1}{l}{\bf{64.9}} \\ 
                                           &OURS*                                                                        & 59.8$\downarrow$\tiny{5.2}   & 77.3$\downarrow$\tiny{4.2}    & 82.8$\downarrow$\tiny{3.3}& 40.6$\downarrow$\tiny{5.4}   & 69.5$\downarrow$\tiny{1.0}   & 86.2$\downarrow$\tiny{1.0}   &90.9$\downarrow$\tiny{1.3}& \multicolumn{1}{l}{63.7$\downarrow$\tiny{1.2}} \\\hline
                                           
Unsup.                                            &BUC \cite{lin2019bottom}                                                                 & 61.1 & 75.1   & 80.0       & 38.0   & 69.2   & 81.1   & 85.8       & 61.9                      \\
One-shot                                           &EUG \cite{wu2018exploit}                                                                     & 62.6 & 74.9   & -       & 42.4 & \bf{72.7}   & 84.1   & -        & 63.2                      \\ 
Intra                                           &UGA \cite{wu2019unsupervised}                                                                 & 59.9 & -      & -       & 40.5 & -      & -      & -        & -                         \\
Fully sup.                                       &Baseline                                                                  & 78.4 & -      & -       & 65.5 & 86.4   & -      & -       & 82.0                        \\ \bottomrule
                 
\end{tabular}
OURS* represents the proposed method under missing annotation. Full sup. denotes fully supervised. Intra indicates Intra-camera supervised.
\end{threeparttable}
\label{table:fine-grained re-id}
\vspace{-0.3cm}
\end{table*}

\subsubsection{Implementation Details}
In this work, an ImageNet \cite{deng2009imagenet} pre-trained ResNet50 network \cite{he2016deep}, in which we replace its last average pooling layer with a {\em d}-dimensional fully connected layer $(d=2048)$, is used as our feature extractor. 
Stochastic gradient descent with a momentum of 0.9 and a batch size of 10 is used to optimize our model. The learning rate is initialized to 0.01 and changed to 0.001 after 10 epochs. We create each batch in a way such that it has a minimum of three pairs of bags and each pair has at least one identity in common.
We train our model end-to-end on two Tesla K80 GPU using Pytorch. We set $k=5$ in Equation 2 for both datasets. 
The number of person images in each training bag is set to a fixed value 100. If the number is greater than that, we randomly select 100 images from the bag and assign the labels of the bag to the selected subset. It may be noted that for WL-DukeV dataset, we split each original person tracklet into 7 parts to increase the number of weakly labeled training samples. To evaluate the performance of our method, the widely used cumulative matching characteristics (CMC) curve and mean average precision (mAP) are used for measurement.

\subsection{Comparison with the Related Methods}
\label{sec:sota}
\subsubsection{Coarse-Grained Person Re-id} 
We compare the performance of our method (MIL and MIL+CPAL) to the existing state-of-the-art multiple instance learning methods - weakly supervised deep detection network (WSDDN) \cite{bilen2016weakly} (section 3.3 of their paper which is relevant for our case), multi-label learning-based hard selection logistic regression (HSLR)  \cite{dong2019single} and soft selection logistic regression (SSLR) \cite{dong2019single} for the task of coarse-grained person re-id. It should be noted that we use the same network architecture for all five methods for fair comparison. From Table \ref{table:coarse-grained re-id}, it can be seen that the proposed {\emph k}-max-mean-pooling based MIL method performs much better than other compared methods. Comparing to WSDDN, the rank-1 accuracy is increased by 9.8\% and 11.0\% for mAP score on WL-MARS dataset. When combining with CPAL (OURS) the recognition performance is further improved. Especially, compared to WSDDN, the rank-1 accuracy and mAP score are improved by $15.2\%$  and $16.8\%$ on WL-MARS dataset, similarly, $10.2\%$ and $9.9\%$ on WL-DukeV dataset.

In this subsection, we also evaluate our method under \emph{missing annotation} scenario. As shown in Table \ref{table:coarse-grained re-id}, we can see that when testing our method under missing annotation situation, for WL-MARS dataset, the rank-1 accuracy and mAP score decrease 0.5\% (78.6\% to 78.1\%) and 4.4\% (47.1\% to 42.7\%), respectively, and for WL-DukeV dataset, it decreases 3.3\% and 3.8\% accordingly. Our method is not very sensitive to missing annotation situation for coarse-grained re-id task. Furthermore, we find that the proposed method with missing annotation still performs significantly better than others with perfect annotation (annotator labels all appeared identities). For example, comparing to HSLR, on WL-MARS dataset, the rank-1 accuracy and mAP score are improved by 8.5\% and 7.3\%, respectively.

\subsubsection{Fine-Grained Person Re-id} 
In Table \ref{table:fine-grained re-id}, we compare our framework against methods which utilize strong labels, as well as other weakly supervised methods for fine-grained person re-id. It can be seen that the proposed {\emph k}-max-mean-pooling-based  MIL  method  performs  much  better  than  most of the other  compared methods and  when  combining with  CPAL  (OURS)  the  recognition  performance  is  further improved. Especially, comparing to HSLR, our method can obtain 8.6\% and 10.2\% improvement for rank-1 accuracy and mAP score respectively, on WL-MARS, and similarly, 8.8\% and 10.2\% improvement on the WL-DukeV dataset. 
The efficacy of using weak labels is strengthened by the improvement over methods which use strong labels, such as EUG (strong labeling: one-shot setting) \cite{wu2018exploit} and UGA (strong labeling: intra-camera supervision) \cite{wu2019unsupervised}. Weak labels also improve performance compared to unsupervised methods such as BUC \cite{lin2019bottom}, with gains of 6.4\% and 8.0\% in rank-5 accuracy and mAP score on WL-MARS dataset, and similarly, 6.1\% and 3.0\% on WL-DukeV dataset. Compared to EUG, the recognition performance is improved from 74.9\% to 81.5\% (6.6\% difference) for rank-5 accuracy on WL-MARS dataset and 84.1\% to 87.2\% (3.1\% difference) on the WL-DukeV dataset.

We evaluate our method under \emph {missing annotation} scenario for fine-grained re-id. As shown in Table \ref{table:fine-grained re-id}, we can see that when testing our method under missing annotation situation, for WL-MARS dataset, the rank-1 accuracy and mAP score decrease 5.2\% and 5.4\%, similarly, 1.0\% and 1.2\% for WL-DukeV dataset. 
We can observe that our results under missing annotation situation are still very competitive compared to others under perfect annotation.
For example, comparing to HSLR, on WL-MARS dataset, the rank-1 accuracy and mAP score are improved by 3.4\% and 4.8\%, similarly, and 7.8\% and 9.0\% on WL-DukeV dataset. Comparing to unsupervised method BUC, we can also obtain better results, especially for the mAP score, our method is 2.6\% and 1.8\% better than that on WL-MARS and WL-DukeV datasets, respectively.

\begin{figure}[t]
\centerline{\includegraphics[width=1\columnwidth]{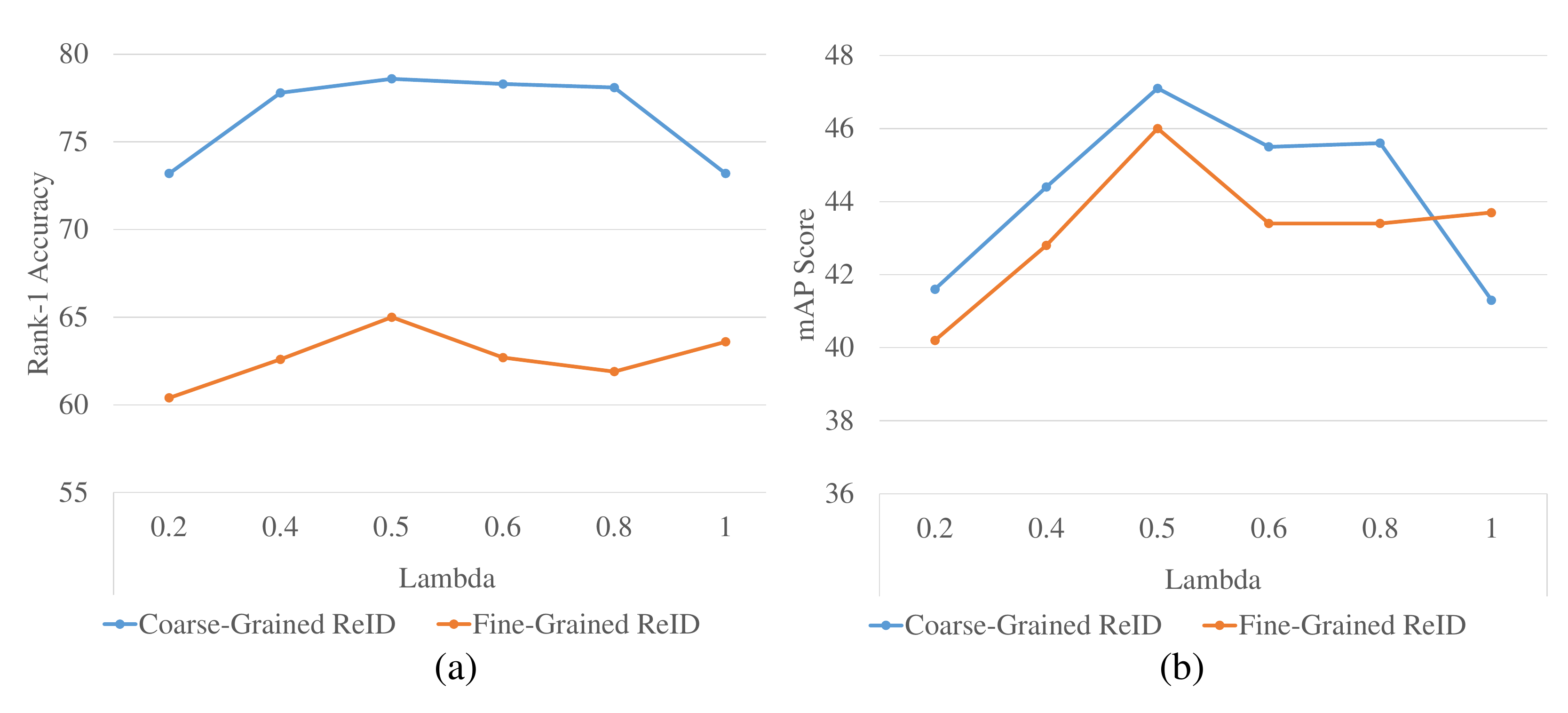}}
\caption{(a) presents the variations in rank-1 accuracy on WL-MARS dataset for coarse and fine-grained re-id tasks by changing parameter $\lambda$. Higher $\lambda$ represents more weights on the MIL and vice versa. (b) presents the variations in mAP score on WL-MARS dataset for both coarse-grained and fine-grained re-id tasks by changing $\lambda$ as discussed in the text.
}
\label{fig:lambda}
\end{figure}

\subsection{Weights Analysis on Loss Functions}
\label{sec:lambda}
In our framework, we jointly optimize MIL and CPAL to learn the weights of the multiple instance attention learning module. In this section, we investigate the relative contributions of the two loss functions to the recognition performance. In order to do that, we perform experiments on WL-MARS dataset, with different values of $\lambda$ (higher value indicates larger weight on MIL), and present the rank-1 accuracy and mAP score on coarse and fine-grained person re-id tasks in Figure \ref{fig:lambda}.

As may be observed from the plot, when $\lambda = 0.5$, the proposed method performs best, {\em i.e.}, both the loss functions have equal weights. Moreover, using only MIL, {\em i.e.}, $\lambda=1.0$, results in a decrease of 5.8\% and 2.3\% in mAP (5.4\% and 1.4\% in rank-1 accuracy) on coarse and fine-grained person re-id tasks, respectively.
This shows that the CPAL introduced in this work has a major effect towards the better performance of our framework.

\subsection{Parameter Analysis}
We adopt a $k$-max-mean-pooling strategy to compute the activation score corresponding to a particular identity in a bag. In this section, we evaluate the effect of varying $k$, which is used in Equation 2. As shown in Table \ref{tab:lambda_param}, the proposed multiple instance attention learning framework is evaluated with four different $k$ values $(k=1,5,10,20)$ on WL-MARS dataset for fine-grained person re-id. It can be seen that when $k=5$, we obtain the best recognition performance 65.0\% for rank-1 accuracy and 46.0\% for mAP score. Comparing to $k=1$ which selects the largest activation for each identity, the performance is improved by 4.0\% and 4.0\% for rank-1 accuracy and mAP score, respectively. We use this value of $k=5$ for all the experiments.


\subsection{Comparison with CV-MIML} \label{sec:cv_miml}
In this section, we compare the proposed framework with CV-MIML \cite{meng2019weakly} that has recently been proposed for weakly supervised person re-id task. Although \cite{meng2019weakly} is presented as a weakly supervised method, it should be noted that it uses a \emph{strongly labeled tracklet} for each identity (one-shot labels) in addition to the weak labels – this is not a true weakly supervised setting and we term it as \emph {semi-weakly supervised}. On the contrary, our method does not require the strong labels and is more in line with the weakly supervised frameworks proposed for object, activity recognition and segmentation \cite{heidarivincheh2019weakly,yu2019temporal,bilen2016weakly,khoreva2017simple,ahn2018learning}. Thus, CV-MIML is not directly applicable to our scenario where one only has access to bags of person images. However, for the sake of comparison, we implemented CV-MIML without the probe set-based MIML loss term $(\mathcal L_p)$ and cross-view bag alignment term $(\mathcal L_{CA})$, since these require the one-shot labels to calculate the cost or the distribution prototype for each class. We refer to this as CV-MIML* and compare it to our method on WL-MARS dataset for coarse-grained re-id task. We also briefly compare our results with the one reported in \cite{meng2019weakly} on Mars dataset.

As shown in Table \ref{tab:cvmiml}, it can be seen that despite the lack of strong labels, our method performs comparably with CV-MIML and completely outperforms its label-free variant CV-MIML* (more than 300\% relative improvement in mAP). In addition, comparing the recognition performance of CV-MIML* and CV-MIML, we find that CV-MIML method relies on strong labels a lot. 

\begin{table}[t]
\centering
\caption{Fine-grained re-id performance comparisons with different parameter $k$ on WL-MARS dataset.}
\setlength{\tabcolsep}{3mm}

\begin{tabular}{llllll}
\toprule
$k$    & Rank-1 & Rank-5 & Rank-10 & Rank-20 & mAP  \\\hline
$k=1$  & 61.0   & 78.5   & 83.4    & 88.0    & 42.0 \\
$k=5$  & \bf 65.0   & \bf 81.5   & \bf 86.1    & \bf 89.7    & \bf 46.0 \\
$k=10$ & 60.6   & 78.0   & 83.2    & 88.4    & 41.7 \\
$k=20$ & 57.5   & 75.9   & 81.2    & 86.0    & 38.4\\
\bottomrule
\end{tabular}
\label{tab:lambda_param}
\end{table}

\begin{table}[t]
\centering
\caption{Coarse-grained re-id performance comparisons with CV-MIML on WL-MARS dataset.}
\setlength{\tabcolsep}{5mm}
\begin{tabular}{lllll}
\toprule
Methods                                             & R1   & R5   & R10  & mAP  \\ \hline
\begin{tabular}[c]{@{}l@{}}CV-MIML*\end{tabular} & 33.3 & 51.3 & 58.5 & 10.7 \\ 
\begin{tabular}[c]{@{}l@{}}CV-MIML \cite{meng2019weakly}\end{tabular}  & 66.8 & 82.0 & 87.2 & 55.1 \\ 
OURS                                                & 78.6 & 90.1 & 93.9 & 47.1 \\ 
\bottomrule
\end{tabular}
\label{tab:cvmiml}
\end{table}

\begin{table}[t]
\centering
\caption{Fine-grained person re-id performance comparisons with tracklet setting.}
\begin{threeparttable}
\setlength{\tabcolsep}{2mm}
\begin{tabular}{llllll}
\toprule
Methods & \multicolumn{1}{l}{Settings}                                                 & Rank-1 & Rank-5 & Rank-10  & mAP  \\\hline
HSLR \cite{dong2019single}   & \multirow{3}{*}{\begin{tabular}[c]{@{}c@{}}Weak\end{tabular}} & 55.4   & 72.8   & 78.6     & 34.7 \\
SSLR \cite{dong2019single}   &                                                                              & 49.0   & 67.9   & 74.0   & 28.7 \\
OURS    &                                                                              & 62.2   & \bf 79.4   & \bf 84.3      & \bf 43.0   \\\hline
BUC \cite{lin2019bottom}    & \multicolumn{1}{l}{None}                                                     & 61.1   & 75.1   & 80.0      & 38.0\\
EUG \cite{wu2018exploit}    & \multicolumn{1}{l}{One-shot}                                                 & \bf 62.6   & 74.9   & -          & 42.4\\
UGA \cite{wu2019unsupervised}     & \multicolumn{1}{l}{Intra}                                                    & 59.9   & -      & -      & 40.5\\
\bottomrule
\end{tabular}
Weak: weak supervision; None: unsupervised; One-shot: a singular labeled tracklet for each identity; Intra denotes intra-camera supervision, in which labels are provided only for samples within an individual camera view.
\end{threeparttable}
\label{tab:tracklet}
\end{table}

\subsection{Evaluation of Multiple Instance Attention Learning with Tracklet Setting}
Our proposed method work with individual frames of the tracklets given in the bag (video). In this section, we perform an ablation study, where we use tracklet features instead of using frame-level features. So, each training sample can be denoted as $(\mathcal X_i,y_i)$ where $\mathcal X_i=\{T_i^1,T_i^2,..,T_i^{m_i}\}$ contains $m_i$ person tracklets and $T_i^k$ is the $k$th tracklet obtained in $i$th video clip, and $y_i$ is a weak label for the bag. Tracklet features are computed by a mean-pooling strategy over the frame features. Table \ref{tab:tracklet} reports the fine-grained person re-id performance on WL-MARS dataset. Even in this setting, our method still performs better than others. Compared to multiple label learning-based HSLR \cite{dong2019single}, we achieve 6.8\% and 8.3\% improvement for rank-1 accuracy and mAP score, respectively. Compared to the state-of-the-art unsupervised BUC \cite{lin2019bottom}, we can also obtain better recognition performance, especially 5\% improvement for mAP score. Moreover, the proposed method is also very competitive compared to those semi-supervised person re-id methods, such as EUG \cite{wu2018exploit} and UGA \cite{wu2019unsupervised}, under tracklet setting. Especially, the mAP score is improved by 0.6\% and 2.5\%, comparing to EUG and UGA, respectively. Next, we present a more practical scenario (Noisy Tracking) where each tracklet may contain more than a singular identity due to the imperfect person tracking in a video clip.

{\emph{Noisy tracking.}}
Assuming correct tracking over the entire duration of a tracklet is a very strong and an unrealistic assumption. Thus, in a practical setting, a tracklet may contain more than a singular identity. Our method obviates this scenario by using frame features. Here, we present the performance using tracklets with noisy tracking. Specifically, we randomly divide the person images in the same bag into 4 parts and regard each of them as a person tracklet that may contain one or multiple person identities. Based on this setting, we compare the fine-grained person re-id performance of the proposed method to a few different methods on WL-MARS dataset. Table \ref{tab:wrong tracking} presents this comparison. Obviously, under noisy tracking setting, the recognition performance declines a lot for all methods comparing to those reported in Table \ref{tab:tracklet}. However, weak supervision-based methods outperform the state-of-the-art unsupervised BUC \cite{lin2019bottom} by a large margin consistently, especially, the proposed method obtains 12.4\% and 11.9\% improvement for rank-1 accuracy and mAP score.

\begin{table}[t]
\centering

\caption{Fine-grained person re-id performance comparisons with noisy tracking.}
\begin{threeparttable}
\setlength{\tabcolsep}{2mm}
\begin{tabular}{llllll}
\toprule
Methods & \multicolumn{1}{l}{Settings}                                                 & Rank-1 & Rank-5 & Rank-10 & mAP  \\\hline
HSLR \cite{dong2019single}    & \multirow{3}{*}{\begin{tabular}[c]{@{}c@{}}Weak\end{tabular}} & 45.0   & 62.5   & 68.9    & 25.2 \\
SSLR \cite{dong2019single}    &                                                                              & 39.0   & 58.1   & 64.2    & 20.4 \\
OURS    &                                                                              & \bf 48.1   & \bf 66.2   & \bf 73.0    & \bf 28.0   \\\hline
BUC \cite{lin2019bottom}    & \multicolumn{1}{l}{None}                                                     & 35.7   & 50.7   & 55.9   & 16.1\\
\bottomrule
\end{tabular}
Weak denotes weak supervision; None denotes unsupervision.
\end{threeparttable}
\label{tab:wrong tracking}
\end{table}

\begin{table}[]
\caption{Ablation studies of the proposed framework on WL-MARS dataset.}
\centering

\begin{tabular}{llllll}
\toprule
\multirow{2}{*}{Settings} & \multirow{2}{*}{Methods} & \multicolumn{4}{c}{WL-MARS}                \\ \cline{3-6} 
                          &                          & R-1 & R-5 & R-10 & mAP  \\ \hline
\multirow{8}{*}{\begin{tabular}[c]{@{}l@{}}Coarse-Grained\\ Re-id\end{tabular}}    
& HSLR                     & 69.6   & 85.9   & 89.8     & 35.4 \\  
                          & HSLR+CPAL                & 74.0     & 87.5   & 93.0     & 42.3 \\ 
                        & SSLR                     & 66.6   & 82.7   & 86.6       & 31.8 \\
  
                          & SSLR+CPAL                & 70.0     & 86.3   & 91.1    & 37.9 \\  
                                                   & WSDDN                    & 63.4   &81.9    & 86.6       & 30.3\\
& WSDDN+CPAL               & 76.4   & 89.7   & 93.4     & 45.9\\
                          & MIL                     & 73.2   & 89.9   & 93.3       & 41.3 \\  
                          & OURS (MIL+CPAL)                & 78.6   & 90.1   & 93.9     & 47.1 \\ \hline
\multirow{8}{*}{\begin{tabular}[c]{@{}l@{}}Fine-Grained\\ Re-id\end{tabular}}   

& HSLR                     & 56.4   & 72.6   & 78.3    & 35.8 \\ 
                          & HSLR+CPAL                & 63.2   & 78.6   & 83.3     & 42.8 \\ 
                          & SSLR                     & 51.9   & 69.3   & 75.7       & 31.2 \\ 
                          & SSLR+CPAL                & 59.3   & 76.2   & 82.2    & 39.4 \\ 
                          & WSDDN &59.2              & 76.4   & 82.4    & 41.7\\
& WSDDN+CPAL &63.6         & 80.3   & 84.0   & 43.1\\
                          & MIL                     & 63.6   & 79.1   & 84.2       & 43.7 \\ 
                          & OURS (MIL+CPAL)                & 65.0     & 81.5   & 86.1    & 46.0   \\ 
                          \bottomrule
\end{tabular}
\label{table:ablation}
\end{table}

\subsection{Ablation Study}
In this section, we conduct ablation studies to evaluate the advantages of our proposed MIL loss and CPAL loss. We validate our methods on WL-MARS dataset under two different tasks - coarse-grained person re-id and fine-grained person re-id. From Table \ref{table:ablation},  we can see that (1) adding CPAL to other methods, such as HSLR+CPAL, SSLR+CPAL and WSDDN+CPAL helps to improve recognition performance by a large margin consistently, such as 4.4\% rank-1 accuracy and 6.9\% mAP score improvement for HSLR-based coarse-grained re-id, and 6.8\% rank-1 accuracy and 7.0\% mAP score improvement for HSLR-based fine-grained re-id; (2) MIL loss performs better than other deep logistic regression-based methods. Comparing to HSLR-based coarse-grained re-id, the rank-1 accuracy is improved from 69.6\% to 73.2\%, and 35.8\% to 43.7\% for mAP score; (3) Combining MIL and CPAL (MIL+CPAL), we can obtain the best recognition performance 78.6\% and 47.1\% for rank-1 accuracy and mAP score on coarse-grained re-id, and 65.0\% and 46.0\% on fine-grained re-id accordingly.

\begin{figure}
\centerline{\includegraphics[width=\columnwidth]{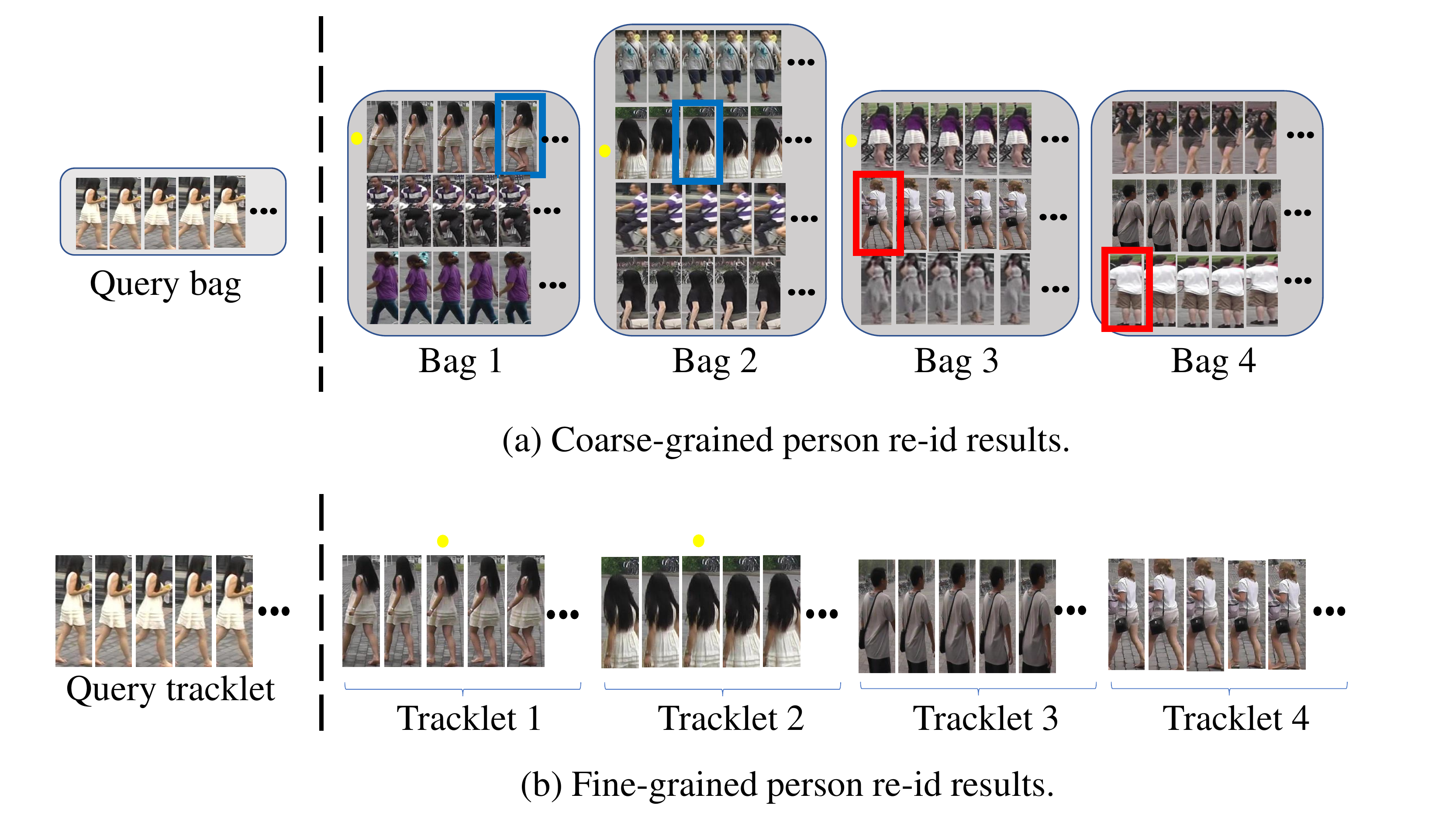}}
\caption{Illustration of coarse-grained person re-id and fine-grained person re-id results on WL-MARS dataset. (a) shows the results of coarse-grained person re-id. It demonstrates 4 retrieved bags (video clips) for a target person. Bounding boxes indicate the the most similar frame in a bag to the target person. Blue and red represent correct and wrong retrieval results. Yellow dots indicate the tracklets with the same identity as the query person. (b) shows the results of fine-grained person re-id. It illustrates 4 retrieved tracklets for a target tracklet.
}
\label{fig:results display}
\end{figure}

\subsection{Matching Examples}
To have better visual understanding, we show some coarse and fine-grained person re-id results achieved by our proposed multiple instance attention learning framework on WL-MARS dataset in Figure \ref{fig:results display}.
Figure \ref{fig:results display}(a) shows the coarse-grained person re-id results. We can see that each query is a bag containing one tracklet with one person identity and 4 returned bags (video clips) are shown in this figure. The bounding boxes indicate the most similar frame in a bag to the query person. Blue and red show the correct and wrong retrieval results, respectively. Yellow dots indicate the tracklets with the same identity as the query person. We find it happens that the most similar frame is wrong, but the retrieval results are correct as shown in Figure \ref{fig:results display}(a): Bag 3. That may explain coarse-grain rank-1 accuracy is better than fine-grained re-id to some extent. Figure \ref{fig:results display}(b) shows us some results of fine-grained person re-id, in which both query and gallery samples are tracklets.

\section{Conclusions}
In this paper, we introduce a novel problem of learning a person re-identification model from videos using weakly labeled data. In the proposed setting, only video-level labels (person identities who appear in the  video) are required, instead of annotating each frame in the video - this significantly reduces the annotation cost. To address this weakly supervised person re-id problem, we propose a multiple instance attention learning framework, in which the video person re-identification task is converted to a multiple instance learning setting, on top of that, a co-person attention mechanism is presented to explore the similarity correlations between videos with common person identities. Extensive experiments on two weakly labeled datasets - WL-MARS and WL-DukeV datasets demonstrate that the proposed framework achieves the state-of-the-art results in the coarse-grained and fine-grained person re-identification tasks. We also validate that the proposed method is promising even when the weak labels are not reliable.

\bibliographystyle{IEEEtran}  
\bibliography{reference.bib}  

\end{document}